\documentclass[10pt,twocolumn,letterpaper]{article}

\usepackage{subfig}
\usepackage{cvpr}
\usepackage{times}
\usepackage{epsfig}
\usepackage{graphicx}
\usepackage{amsmath}
\usepackage{amssymb}

\usepackage[pagebackref=true,breaklinks=true,letterpaper=true,colorlinks,bookmarks=false]{hyperref}

\cvprfinalcopy 

\def\cvprPaperID{8} 

\ifcvprfinal\fi
\begin{document}

\title{Facial Expression Recognition Using Enhanced Deep 3D Convolutional Neural Networks}

\author{Behzad Hasani and Mohammad H. Mahoor\\
Department of Electrical and Computer Engineering	\\
 University of Denver, Denver, CO\\
{\tt\small behzad.hasani@du.edu and mmahoor@du.edu}
}

\maketitle

\begin{abstract}
   
   Deep Neural Networks (DNNs) have shown to outperform traditional methods in various visual recognition tasks including Facial Expression Recognition (FER). In spite of efforts made to improve the accuracy of FER systems using DNN, existing methods still are not generalizable enough in practical applications. This paper proposes a 3D Convolutional Neural Network method for FER in videos. This new network architecture consists of 3D Inception-ResNet layers followed by an LSTM unit that together extracts the spatial relations within facial images as well as the temporal relations between different frames in the video. Facial landmark points are also used as inputs to our network which emphasize on the importance of facial components rather than the facial regions that may not contribute significantly to generating facial expressions. Our proposed method is evaluated using four publicly available databases in subject-independent and cross-database tasks and outperforms state-of-the-art methods.
\end{abstract}

\section{Introduction}

Facial expressions are one of the most important nonverbal channels for expressing internal emotions and intentions. Ekman \emph{et al.} \cite{c5} defined six expressions (viz.  anger, disgust, fear, happiness, sadness, and surprise) as basic emotional expressions which are universal among human beings. Automated Facial Expression Recognition (FER) has been a topic of study for decades. Although there have been many breakthroughs in developing automatic FER systems, majority of the existing methods either show undesirable performance in practical applications or lack generalization due to the controlled condition in which they are developed \cite{c2}. 

The FER problem becomes even more difficult when we recognize expressions in videos. 
Facial expressions have a dynamic pattern that can be divided into three phases: onset, peak and offset, where the onset describes the beginning of the expression, the peak (aka apex) describes the maximum intensity of the expression and the offset describes the moment when the expression vanishes. Most of the times, the entire event of facial expression from the onset to the offset is very quick, which makes the process of expression recognition very challenging \cite{c3}.

\begin{figure}[!tbp]
	\begin{center}
		\includegraphics[width=\linewidth]{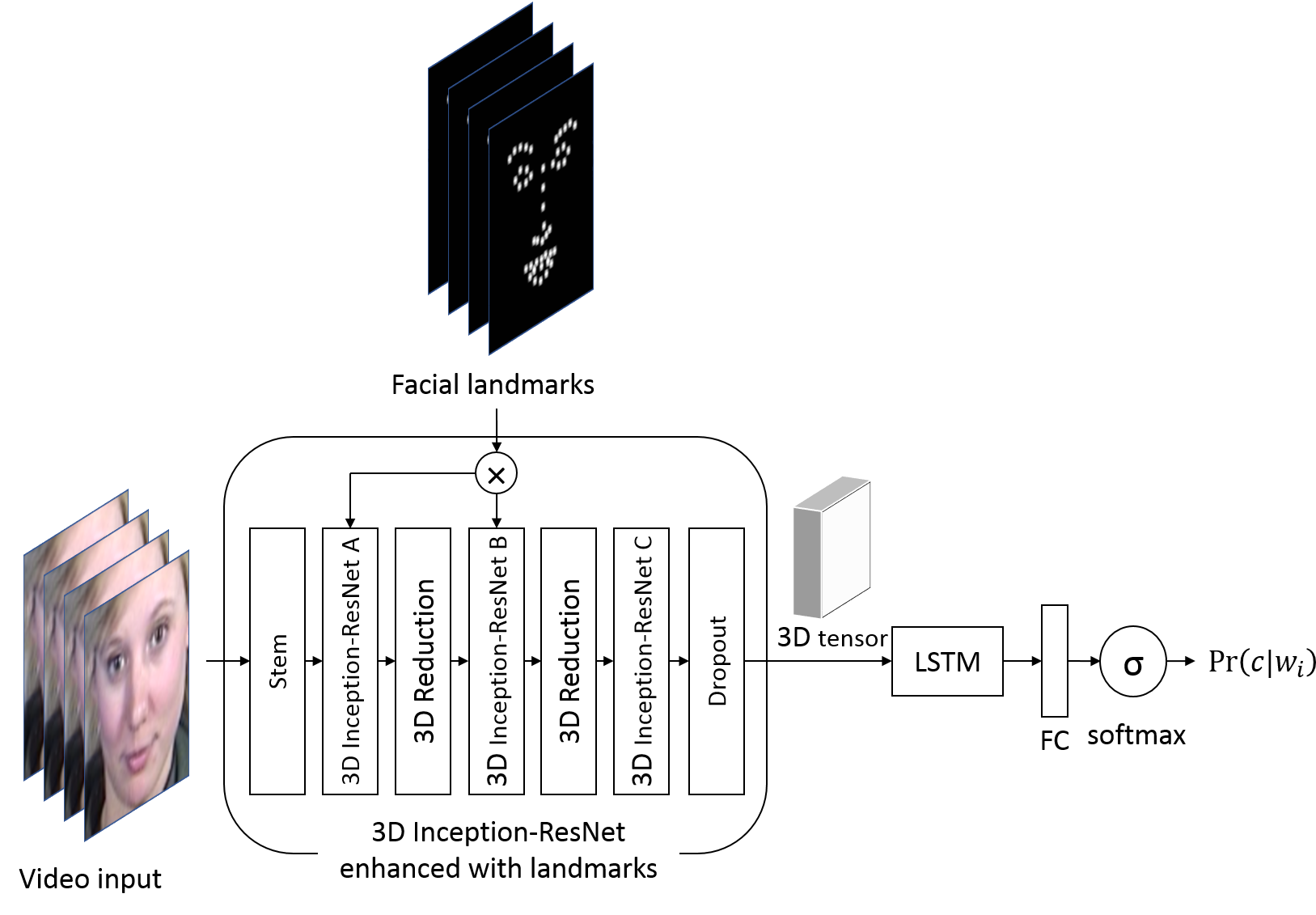}
	\end{center}
	\caption{Proposed method}
	\label{fig:method}
\end{figure}

Many methods have been proposed for automated facial expression recognition. Most of the traditional approaches mainly consider still images independently while ignore the temporal relations of the consecutive frames in a sequence which are essential for recognizing subtle changes in the appearance of facial images especially in transiting frames between emotions. Recently, with the help of Deep Neural Networks (DNNs), more promising results are reported in the field~\cite{c4,c53}. 
  While in traditional approaches engineered features are used to train classifiers, DNNs  have the ability  to extract more discriminative features which yield in a better interpretation of the texture of human face in visual data. 
  
  One of the problems in FER is that training neural networks is significantly more difficult as most of the existing databases have a small number of images or video sequences for certain emotions~\cite{c53}. Also, most of these databases contain still images that are unrelated to each other (instead of having consecutive frames of exhibiting the expression from onset to offset)  which makes the task of sequential image labeling more difficult.

In this paper, we propose a method which extracts temporal relations of consecutive frames in a video sequence using 3D convolutional networks and Long Short-Term Memory (LSTM). Furthermore, we extract and incorporate facial landmarks in our proposed method that emphasize on more expressive facial components which improve the recognition of subtle changes in the facial expressions in a sequence (Figure~\ref{fig:method}). We evaluate our proposed method using four well-known facial expression databases (CK+, MMI, FERA, and DISFA) in order to classify the expressions. Furthermore, we examine the ability of our method in recognition of facial expressions in cross-database classification tasks.

The remainder of the paper is organized as follows: Section \ref{sec:2}  provides an overview of the related work in this field. Section \ref{sec:3} explains the network proposed in this research. Experimental results and their analysis are presented in Section~\ref{sec:4} and finally the paper is concluded in Section \ref{sec:6}. 

\section{Related work}\label{sec:2}

Traditionally, algorithms for  automated  facial expression recognition consist of three main modules,  viz.   registration,  feature extraction,  and classification. Detailed survey of different approaches in each of these steps can be found in \cite{c9}. Conventional algorithms for affective computing from faces use engineered features such as Local  Binary  Patterns (LBP) \cite{c2}, Histogram of Oriented Gradients  (HOG)  \cite{c14}, Local Phase Quantization (LPQ) \cite{c13},  Histogram of Optical Flow \cite{c15}, facial landmarks~\cite{c11,c12}, and PCA-based methods \cite{c54}. Since the majority of these features are hand-crafted for their specific application of recognition, they often lack required generalizability in cases where there is high variation in lighting, views, resolution, subjects' ethnicity, etc.


One of the  effective approaches for achieving better recognition rates for sequence labeling task is  to extract the temporal relations of frames in a sequence. Extracting these temporal relations has been studied using traditional methods in the past. Examples of these attempts are Hidden Markov
Models~\cite{c24,c25,c28} (which combine
temporal information and apply segmentation on videos), Spatio-Temporal Hidden Markov Models (ST-HMM) by coupling
S-HMM and T-HMM~\cite{C29}, Dynamic Bayesian Networks (DBN)~\cite{c17,c26} associated with a
multi-sensory information fusion strategy, Bayesian temporal models~\cite{c27} to capture the dynamic facial expression transition, and Conditional Random Fields (CRFs)~\cite{7869885,behzad2017, c33,c30} and their extensions such as Latent-Dynamic
Conditional Random Fields (LD-CRFs) and Hidden Conditional Random
Fields (HCRFs)~\cite{c31}.

In recent years, ``Convolutional Neural Networks" (CNNs) have become the most popular  approach  among  researchers in the field.  AlexNet \cite{c21} is based on the traditional CNN layered architecture which consists of several convolution layers followed by max-pooling layers and Rectified Linear Units (ReLUs). 
Szegedy \emph{et al.} \cite{c18} introduced GoogLeNet which is composed of  multiple  ``Inception"  layers. Inception applies several convolutions on the feature map in different scales. Mollahosseini \emph{et al.} \cite{c4,c53} have used the Inception layer for the task of facial expression recognition and achieved state-of-the-art results. Following the success of Inception layers, several variations of them have been proposed~\cite{c20,c19}. Moreover, Inception layer is combined with residual unit introduced by He \emph{et al.}  \cite{c6} and it shows that the resulting architecture accelerates the training of Inception networks significantly~\cite{c23}.

One of the major restrictions of ordinary Convolutional Neural Networks is that they only extract spatial relations of the input data while ignore the temporal relations of them if they are part of a sequenced data. To overcome this problem, 3D Convolutional Neural Networks (3D-CNNs) have been proposed. 3D-CNNs slide over the temporal dimension of the input data as well as the spatial dimension enabling the network to extract feature maps containing temporal information which is essential for sequence labeling tasks. Song \emph{et al.}~\cite{song2016deep} have used 3D-CNNs for 3D object detection task. Molchanov \emph{et al.}~\cite{molchanov2016online} have proposed a recurrent 3D-CNN for dynamic hand gesture recognition and Fan \emph{et al.}~\cite{Fan:2016} won the EmotiW 2016 challenge by cascading 3D-CNNs with LSTMs.

Traditional Recurrent Neural Networks (RNNs) can learn temporal dynamics by mapping input sequences to a sequence of hidden states, and also mapping the hidden states to outputs~\cite{donahue2015long}. Although RNNs have shown promising performance on various tasks, it is not easy for them to learn long-term sequences. This is mainly due to the vanishing/exploding gradients problem~\cite{C60} which can be solved by having a memory for remembering and forgetting the previous states. LSTMs~\cite{C60} provide such memory and can memorize the context information for long periods of time. LSTM modules have three gates: 1) the input gate $(i)$ 2) the forget gate $(f)$ and 3) the output gate $(o)$ which overwrite, keep, or retrieve the memory cell $c$ respectively at the timestep $t$. Letting $ \sigma(x)=(1+\exp(-x))^{-1} $ be the $sigmoid$ function and $ \phi(x) = \frac{\exp(x)-\exp(-x)}{\exp(x)+\exp(-x)}= 2 \sigma (2x)-1$ be the $ hyperbolic$ $tangent $ function. Letting $x$, $h$, $c$, $W$, and $b$ be the input, output, cell state, parameter matrix, and parameter vector respectively. The LSTM updates for the timestep $t$ given inputs $x_t$, $h_{t-1}$, and $c_{t-1}$ are as follows:

\begin{equation}
\begin{split}
&f_t = \sigma (W_f\cdot[h_{t-1},x_t] + b_f) \\
&i_t = \sigma (W_i\cdot[h_{t-1},x_t] + b_i) \\
&o_t = \sigma (W_o\cdot[h_{t-1},x_t] + b_o) \\
&g_t =  \phi (W_C\cdot[h_{t-1},x_t] + b_C)\\
&C_t = f_t \ast C_{t-1} + i_t \ast g_t \\
&h_t = o_t \ast \phi ( C_t )
\end{split}
\label{eq:LSTM}
\end{equation}

Several works have used LSTMs for  the task of sequence labeling. Byeon \emph{et al.}~\cite{byeon2015scene} proposed an LSTM-based network applying LSTMs in four direction sliding windows and achieved impressive results. Fan \emph{et al.}~\cite{Fan:2016} cascaded 2D-CNN with LSTMs and combined the feature map with 3D-CNNs for facial expression recognition. Donahue \etal~\cite{donahue2015long} proposed  Long-term  Recurrent  Convolutional  Network (LRCN) by combining CNNs and LSTMs which is both spatially and temporally deep and has the flexibility to be applied to different vision tasks involving sequential inputs and outputs.


\section{Proposed method}\label{sec:3}
While Inception and ResNet have shown remarkable results in FER~\cite{behzad2017,c18}, these methods do not extract the temporal relations of the input data. Therefore, we propose a 3D Inception-ResNet architecture  to address this issue.  Our proposed method, extracts both spatial and temporal features of the sequences in an end-to-end neural network. Another component of our method is incorporating facial landmarks in an automated manner during training in the proposed neural network. These facial landmarks help the network to pay more attention to the important facial components in the feature maps which results in a more accurate recognition. The final part of our proposed method is an LSTM unit which takes the enhanced feature map resulted from the 3D Inception-ResNet (3DIR) layer as an input and extracts the temporal information from it. The LSTM unit is followed by a fully-connected layer associated with a softmax activation function. In the following, we explain each of the aforementioned units in detail.

\subsection{3D Inception-ResNet (3DIR)}
We propose 3D version of Inception-ResNet network which is  slightly shallower  than the original Inception-ResNet network proposed in~\cite{c23}. This network is the result of investigating several variations of Inception-ResNet module and achieves better recognition rates comparing to our other attempts in several databases.

\begin{figure}[!tbp]
	\begin{center}
	\includegraphics[width=\linewidth]{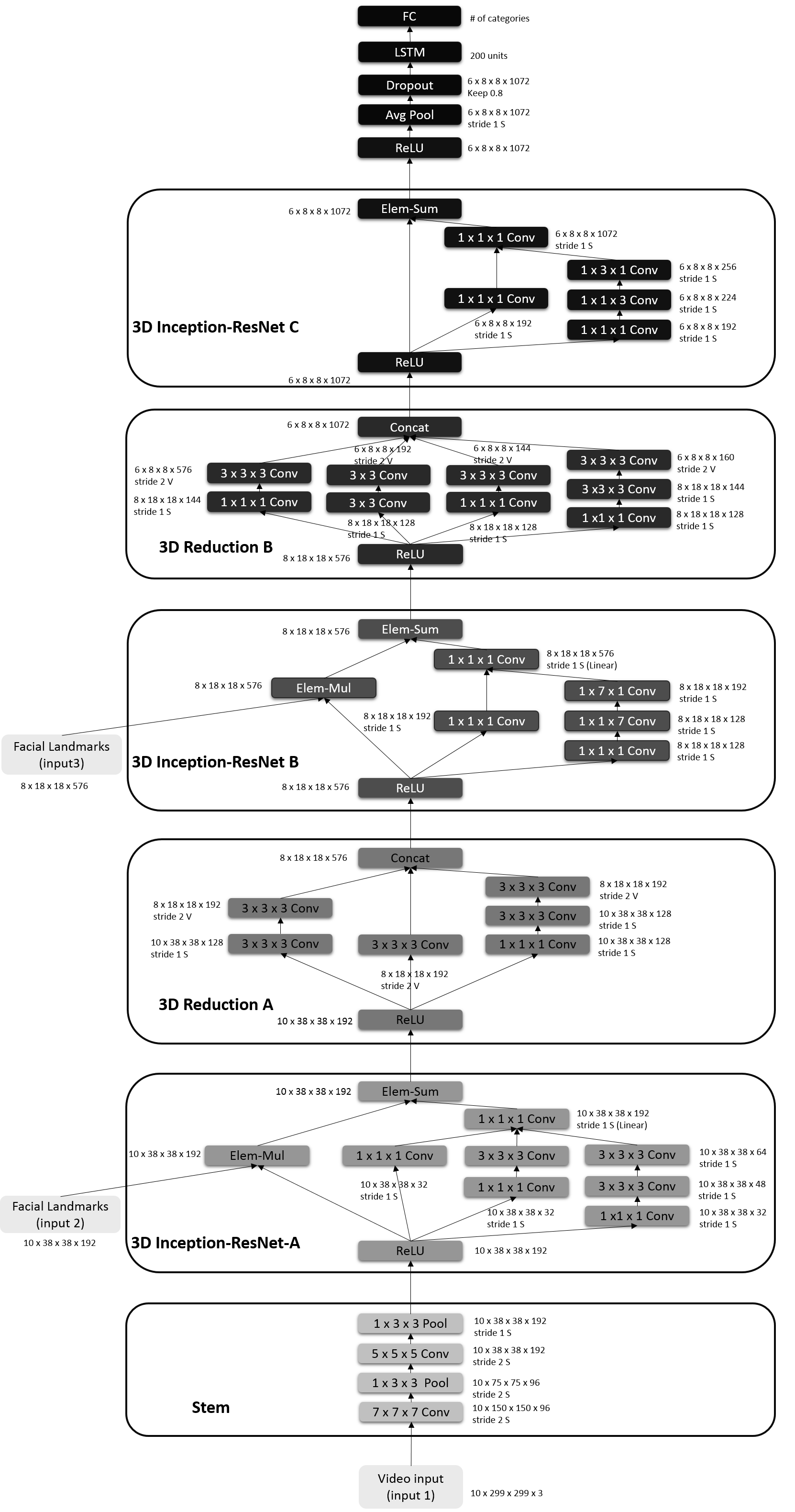}
	\end{center}
	\caption{Network architecture. The ``V" and ``S" marked layers represent ``Valid" and ``Same" paddings respectively. The size of the output tensor is provided next to each layer.}
	\label{fig:Net}
\end{figure}

Figure \ref{fig:Net} shows the structure of our 3D Inception-ResNet network. The input videos with the size $10\times299\times299\times3$ ($10$ frames, $299\times299$ frame size and $3$ color channels) are followed by the ``stem" layer. Afterwards, stem is followed by 3DIR-A, Reduction-A (which reduces the grid size from $38\times38$ to $18\times18$), 3DIR-B, Reduction-B (which reduces the grid size from $18\times18$ to $8\times8$), 3DIR-C, Average Pooling, Dropout, and a fully-connected layer respectively. In Figure \ref{fig:Net}, detailed specification of each layer is provided. Various filter sizes, paddings, strides, and activations have been investigated and the one that had the best performance is presented in this paper.  

We should mention that all convolution layers (except the ones that are indicated as ``Linear" in Figure \ref{fig:Net}) are followed by an ReLU \cite{c21} activation function to avoid the vanishing gradient problem.

\subsection{Facial landmarks}

As mentioned before, the main reason we use facial landmarks in our network is to differentiate between the importance of main facial components (such as eyebrows, lip corners, eyes, etc.) and other parts of the face which are less expressive of facial expressions. As oppose to general object recognition task, in FER, we have the advantage of extracting facial landmarks and using this information to improve the recognition rate. In a similar approach, Jaiswal \etal~\cite{jaiswal2016deep} proposed incorporation of binary masks around different parts of the face in order to encode the shape of different face components. However, in this work authors perform AU recognition by using CNN as a feature extractor for training  Bi-directional Long Short-Term Memory while in our approach, we preserve the temporal order of the frames throughout the network and train CNN and LSTMs simultaneously in an end-to-end network. We incorporate the facial landmarks by replacing the shortcut in residual unit on original ResNet with element-wise multiplication of facial landmarks and the input tensor of the residual unit (Figures~\ref{fig:method} and~\ref{fig:Net}).

In order to extract the facial landmarks, OpenCV face recognition is used to obtain bounding boxes of the faces. A face alignment algorithm via regression
local binary features~\cite{ren2014face,LequanYu2016} was used to extract 66 facial landmark points. The facial landmark localization technique was trained using the annotations provided from the 300W competition~\cite{sagonas2015300,sagonas2013semi}.

After detecting and saving the facial landmarks for all of the databases, the facial landmark filters are generated for each sequence automatically during training phase. Given the facial landmarks for each frame of a sequence, we initially resize all of the images in the sequence to their corresponding filter size in the network. Afterwards, we assign weights to all of the pixels in a frame of a sequence based on their distances to the detected landmarks. The closer a pixel is to a facial landmark, the greater weight is assigned to that pixel. After investigating several distance measures, we concluded that $Manhattan$ distance with a linear weight function results in a better recognition rate in various databases. The Manhattan distance between two items is the sum of the differences of their corresponding components (in this case two components). 

\begin{figure}[!tbp]
	\centering
	\subfloat[Landmarks]{\includegraphics[width=.15\textwidth]{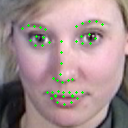}\label{fig:landmark}}
	\hspace{1em}
	\subfloat[Generated filter]{\includegraphics[width=.15\textwidth]{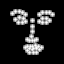}\label{fig:filter}}
	\caption{Sample image from MMI database (left) and its corresponding filter in the network (right). Best in color.}
	\label{fig:facial_landmark_and_filter}
\end{figure}

The weight function that we defined to assign the weight values to their corresponding feature is a simple linear function of the Manhattan distance defined as follows:

\begin{equation}
\omega(L,P)= 1- 0.1\cdot d_{M(L,P)} 
\label{eq:weight_function}
\end{equation}
where $d_{M(L,P)}$ is the Manhattan distance between the facial landmark $L$ and pixel $P$. Therefore, places in which facial landmarks are located will have the highest value and their surrounding pixels will have lower weights proportional to their distance from the corresponding facial landmark. In order to avoid overlapping between two adjacent facial landmarks, we define a $7\times7$ window around each facial landmark and apply the weight function for these 49 pixels for each landmark separately. Figure~\ref{fig:facial_landmark_and_filter} shows an example of facial image from MMI database and its corresponding facial landmark filter in the network. We do not incorporate the facial landmarks with the third 3D Inception-ResNet module since the resulting feature map size at this stage becomes very small for calculating facial landmark filter.

Incorporating facial landmarks in our network replaces the shortcut in original ResNets~\cite{c22} with the element-wise multiplication of the weight function $\omega$ and input layer $x_l$ as follows:

\begin{equation}
\begin{split}
&y_l = \omega(L,P) \circ x_l + F (x_l, W_l)\\
& x_{l+1} = f(y_l)
\end{split}
\label{eq:residual_formula}
\end{equation}
where $x_l$ and $x_{l+1}$ are input and output of the $l$-th layer, $\circ$ is Hadamard product symbol, $F$ is a residual function (in our case Inception layer convolutions), and $f$ is an activation function.

\subsection{Long Short-Term Memory unit}

As explained earlier, to capture the temporal relations of the resulted feature map from 3DIR and take these relations into account by the time of classifying the sequences in the softmax layer, we used an LSTM unit as it is shown in Figure~\ref{fig:Net}. Using the LSTM unit makes perfect sense since the resulted feature map from the 3DIR unit contains the time notion of the sequences within the feature map. Therefore, vectorizing the resulting feature map of 3DIR on its sequence dimension, will provide the required sequenced input for the LSTM unit. While other still image LSTM-based methods, a vectorized non-sequenced feature map (which obviously does not contain any time notion) is fed to the LSTM unit, our method saves the time order of the input sequences and passes this feature map to the LSTM unit. We investigated that 200 hidden units for the LSTM unit is a reasonable amount for the task of FER (Figure~\ref{fig:Net}).

The  proposed  network  was  implemented  using a combination of TensorFlow~\cite{abadi2016tensorflow} and TFlearn~\cite{tflearn2016} toolboxes on NVIDIA Tesla K40 GPUs. In the training phase we used asynchronous stochastic gradient descent with momentum of  0.9, weight decay of 0.0001, and learning rate of 0.01. We used categorical cross entropy as our loss function and accuracy as our evaluation metric.

\section{Experiments and results}\label{sec:4}
In this section, we briefly review the databases we used for evaluating our method. We then report the results of our experiments using these databases and compare the results with the state of the arts.

\subsection{Face databases}

Since our method is designed mainly for classifying sequences of inputs, databases that contain only independent unrelated still images of facial expressions such as MultiPie~\cite{c55} , SFEW~\cite{c56} , FER2013~\cite{c57} cannot be examined by our method. We evaluate our proposed method on MMI~\cite{c37}, extended CK+~\cite{c38}, GEMEP-FERA~\cite{c39}, and DISFA~\cite{c59} which contain videos of annotated facial expressions. In the following, we briefly review the contents of these databases.

{\bf MMI:} The  MMI   \cite{c37}  database  contains  more  than  20 subjects, ranging in age from 19 to 62, with different ethnicities (European, Asian, or South American). In MMI, the subjects' facial expressions start from the neutral state to the apex of one of the six basic facial expressions and then returns to the neutral state again. Subjects were instructed to display 79 series of facial expressions,  six  of  which  are  prototypic  emotions (angry, disgust, fear, happy, sad, and surprise).   We  extracted static frames from each sequence, which resulted in 11,500 images. Afterwards, we divided videos into sequences of ten frames to shape the input tensor for our network.

{\bf CK+:} The extended Cohn-Kanade database (CK+) \cite{c38} contains 593 videos from 123 subjects. However, only 327 sequences from 118 subjects contain facial expression labels. Sequences in this database start from the neutral state and end at the apex of one of the six basic expressions (angry, contempt, disgust, fear, happy, sad, and surprise). CK+ primarily contains frontal face poses only. In order to make the database compatible with our network, we consider the last ten frames of each sequence as an input sequence in our network.

{\bf FERA:} The GEMEP-FERA database \cite{c39} is a subset of the  GEMEP  corpus  used  as  database  for  the  FERA  2011 challenge~\cite{c41} developed by the Geneva Emotion Research Group at the University of Geneva. This database contains 87 image sequences of 7 subjects. Each subject shows facial expressions of the emotion categories: Anger, Fear, Joy, Relief, and Sadness. Head pose is primarily frontal with relatively fast movements. Each video is annotated with AUs and holistic expressions. By extracting static frames from the sequences, we obtained around 7,000 images. We divided the these emotion videos into sequences of ten frames to shape the input tensor for our network.

\textbf{DISFA:} Denver Intensity of Spontaneous Facial Actions (DISFA) database~\cite{c59} is one of a few naturalistic databases that have been FACS coded by AU intensity values. This database consists of 27 subjects. The subjects are asked to watch YouTube videos while their spontaneous facial expressions are recorded. Twelve AUs are coded for each frame and AU intensities are on a six-point scale between 0-5, where 0 denotes the absence of the AU, and 5 represents maximum intensity. As DISFA is not emotion-specified coded, we used EMFACS system~\cite{friesen1983emfacs} to convert AU FACS codes to seven expressions (angry, disgust, fear, happy, neutral, sad, and surprise) which resulted in around 89,000 images in which the majority have neutral expressions. Same as other databases, we divided the videos of emotions into sequences of ten frames to shape the input tensor for our network.

\subsection{Results}

As mentioned earlier, after detecting faces we extract 66 facial landmark points by a face alignment algorithm via regression
local binary features. Afterwards, we resize the faces to $299\times299$ pixels. One of the reasons why we choose large image size as input is the fact that larger images and sequences will enable us to have deeper networks and extract more abstract features from sequences. All of the networks have the same settings (shown in Figure~\ref{fig:Net} in detail) and are trained from scratch for each database separately.

We evaluate the accuracy of our proposed method with two different sets of experiments: ``subject-independent" and ``cross-database" evaluations. 

\subsubsection{Subject-independent task}

In the subject-independent  task,  each database  is  split  into training and  validation  sets  in  a  strict  subject  independent manner. In all databases, we report the results using the 5-fold cross-validation technique and then averaging the recognition rates over five folds. For each database and each fold, we trained our proposed network entirely from scratch with the aforementioned settings. Table~\ref{subject_ind_table}  shows  the  recognition rates achieved on each database in the subject-independent case and compares the results with the state-of-the-art methods. In order to compare the impact of incorporating facial landmarks, we also provide the results of our network while the landmark multiplication unit is removed and replaced with a simple shortcut between the input and output of the residual unit. In this case, we randomly select 20 percent of the subjects as the test set and report the results on those subjects. Table~\ref{subject_ind_table} also  provides the recognition rates of the traditional 2D Inception-ResNet from~\cite{behzad2017} which does not contain facial landmarks and the LSTM unit (DISFA is not experimented in this study).

\begin{table*}[]
	\begin{center}
	\begin{tabular}{l|c|c|c|c|}
		\cline{2-5}
		                            & state-of-the-art methods                                                                                                                                                       & \begin{tabular}[c]{@{}c@{}}2D\\ Inception-ResNet\end{tabular}     & \begin{tabular}[c]{@{}c@{}}3D\\ Inception-ResNet\end{tabular}            & \begin{tabular}[c]{@{}c@{}}3D\\ Inception-ResNet\\ + \\landmarks\end{tabular} \\ \hline
		\multicolumn{1}{|l|}{CK+}   &\begin{tabular}[c]{@{}c@{}}84.1~\cite{c42}, 84.4~\cite{c45}, 88.5~\cite{c46}, 92.0~\cite{c47},\\ 93.2~\cite{c4}, 92.4~\cite{c48}, 93.6~\cite{c44}\end{tabular} & 85.77                                                             &  89.50                                                                   & 93.21$\pm$2.32                                                                \\ \hline
		\multicolumn{1}{|l|}{MMI}   &\begin{tabular}[c]{@{}c@{}}63.4~\cite{c48}, 75.12~\cite{c51}, 74.7~\cite{c47}, 79.8~\cite{c46},\\  86.7~\cite{c2}, 78.51~\cite{c54}   \end{tabular}                       & 55.83                                                             &  67.50                                                                   & 77.50$\pm$1.76                                                                \\ \hline
		\multicolumn{1}{|l|}{FERA}  &\begin{tabular}[c]{@{}c@{}}    \end{tabular}  56.1~\cite{c48}, 55.6~\cite{c49},  76.7~\cite{c4}                                                                           & 49.64                                                             &  67.74                                                                   &  77.42$\pm$3.67                                                              \\ \hline
		\multicolumn{1}{|l|}{DISFA} &55.0~\cite{c4}                                                                                                                                                           & -                                                                 &  51.35                                                                   &  58.00$\pm$5.77                                                             \\ \hline
	\end{tabular}
	\end{center}
	\caption{Recognition rates (\%) in subject-independent task}
	\label{subject_ind_table}
\end{table*}

Comparing the recognition rates of the 3D and 2D Inception-ResNets in Table~\ref{subject_ind_table}, shows that the sequential processing of facial expressions considerably enhances the recognition rate. This improvement is more apparent in MMI and FERA databases. Incorporating landmarks in the network is proposed to emphasize  on more important facial changes over time. Since changes in the lips or eyes are much more expressive than the changes in other components such as the cheeks, we utilize facial landmarks to enhance these temporal changes in the network flow. 

The ``3D Inception-ResNet with landmarks" column in Table~\ref{subject_ind_table} shows the impact of this enhancement in different databases. It can be seen that compared with other networks, there is a considerable improvement in recognition rates especially in FERA and MMI databases. The results on DISFA, however, show higher fluctuations over different folds which can be in part due to the abundance of inactive frames in this database which causes confusion in recognizing different expressions. Therefore, the folds that contain more neutral faces, would show lower recognition rates.

Comparing to other state-of-the-art works, our method outperforms others in FERA and DISFA databases while achieves comparable results in CK+ and MMI databases (Table~\ref{subject_ind_table}). Most of these works use traditional approaches including hand-crafted features  tuned for that specific database, while our network's settings are the same for all databases. Also, due to the limited number of samples in these databases, it is difficult to properly train a deep neural network and avoid the overfitting problem. For these reasons and in order to have a better understanding about our proposed method, we also experimented the cross-database task.

\begin{figure*}[!tbp]
	\centering
	\subfloat[CK+]{\includegraphics[width=.30\textwidth]{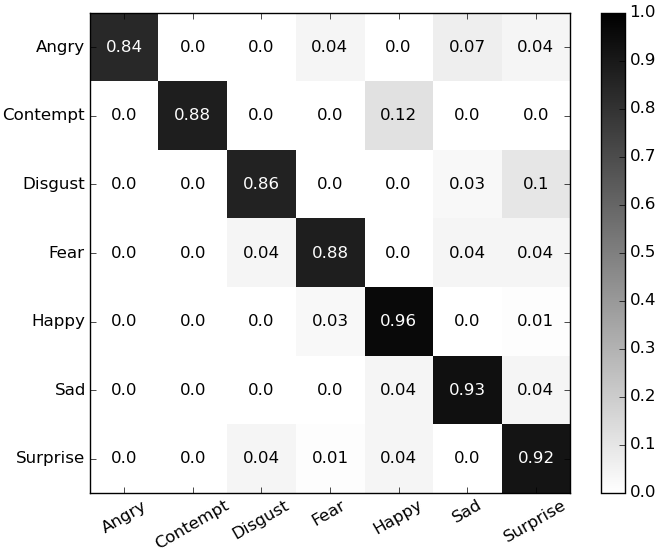}\label{fig:CK+_CM}}
	\hspace{2em}
	\subfloat[MMI]{\includegraphics[width=.30\textwidth]{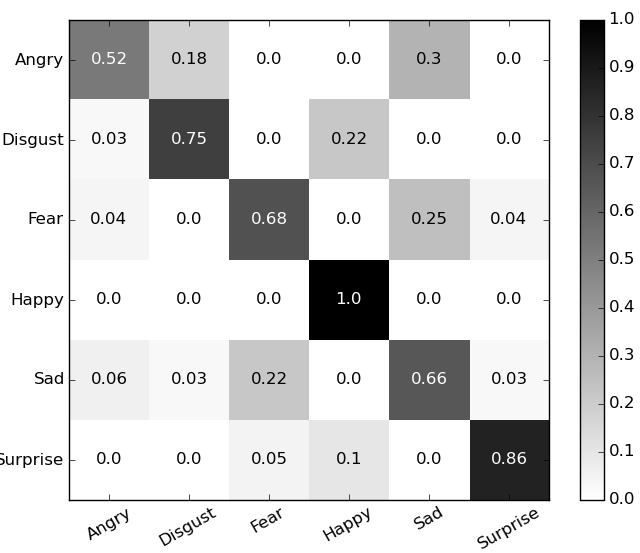}\label{fig:MMI_CM}}
	\hspace{2em}\\
	\subfloat[FERA]{\includegraphics[width=.30\textwidth]{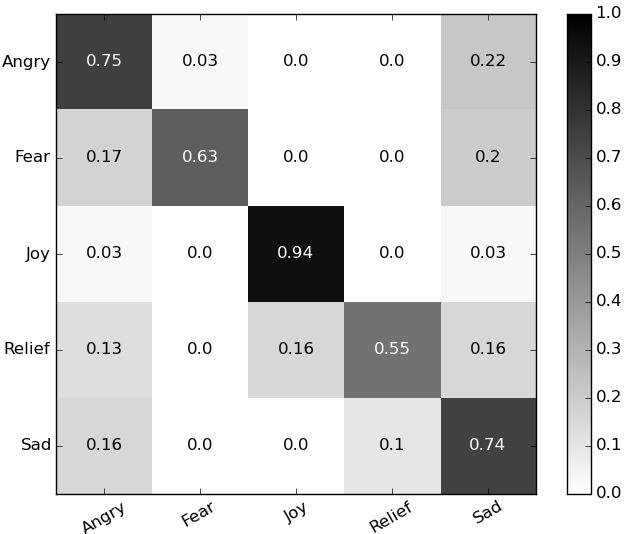}\label{fig:FERA_CM}}
	\hspace{2em}
	\subfloat[DISFA]{\includegraphics[width=.30\textwidth]{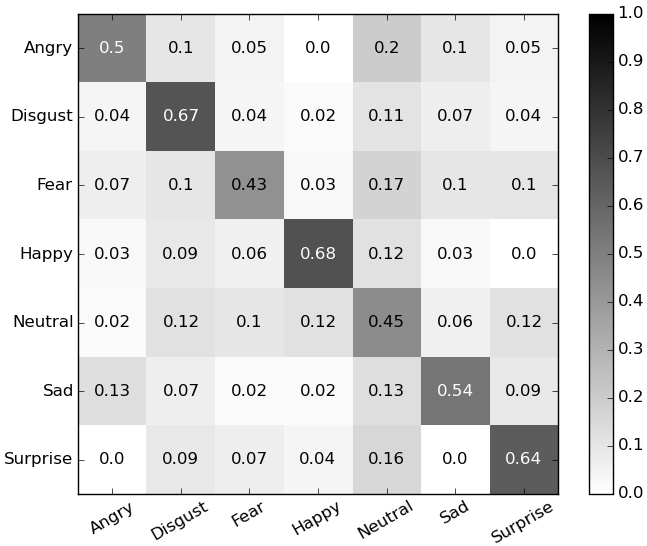}\label{fig:DISFA_CM}}
	\caption{Confusion matrices of 3D Inception-ResNet with landmarks for subject-independent task}
	\label{fig:confusion_matrices}
\end{figure*}

Figure~\ref{fig:confusion_matrices} shows the resulting confusion matrices of our 3D Inception-ResNet with incorporating landmarks on different databases over the 5 folds. On CK+ (Figure~\ref{fig:CK+_CM}), it can be seen that very high recognition rates have been achieved. The recognition rates of happiness, sadness, and surprise are higher than those of other expressions. The highest confusion occurred between the happiness and contempt expressions which can be caused from the low number of contempt sequences in this database (only 18 sequences). On MMI (Figure~\ref{fig:MMI_CM}), a perfect recognition is achieved for the happy expression. It can be seen that there is a high confusion between the sad and fear expressions as well as the angry and sad expressions. Considering the fact that MMI is a highly imbalanced dataset, these confusions are reasonable. On FERA (Figure~\ref{fig:FERA_CM}), the highest and the lowest recognition rates belong to joy and relief respectively. The relief category in this database has some similarities with other categories especially with joy. These similarities make the classification so difficult even for humans. Despite these challenges, our method has performed well on all of the categories and outperforms state of the arts. On DISFA (Figure~\ref{fig:DISFA_CM}), we can see the highest confusion rate compared with other databases. As mentioned earlier, this database contains long inactive frames, which means that the number of neutral sequences is considerably higher than other categories. This imbalanced training data has made the network to be biased toward the neutral category and therefore we can observe a high confusion rate between the neutral expression and other categories in this database. Despite the low number of angry and sad sequences in this database, our method has been able achieve satisfying recognition rates in these categories.   

\subsubsection{Cross-database task}

In the cross-database task, for testing each database, that database is entirely  used for testing the network and the rest of the databases are used to train the network. The same network architecture as subject-independent task (Figure~\ref{fig:Net}) was used for this task. Table \ref{cross_database_table} shows the recognition rate achieved on each database in the cross-database case and it also compares the results with other state-of-the-art methods. 
It can be seen that our method outperforms the state-of-the-art results in CK+, FERA, and DISFA databases. On MMI, our method does not show improvements comparing to others (\eg ~\cite{c44}). However, authors in~\cite{c44} trained their classifier only with CK+ database while our method uses instances from two additional databases (DISFA and FERA) with completely different settings and subjects which add significant amount of ambiguity in the training phase.

\begin{table}[!b]
	\begin{center}
	\begin{tabular}{l|c|c|c|c|}
		\cline{2-3}
		                            & state-of-the-art methods                                                                                                                           & \begin{tabular}[c]{@{}c@{}}3D\\ Inception-ResNet\\ + \\ landmarks\end{tabular}   \\ \hline
		\multicolumn{1}{|l|}{CK+}   & \begin{tabular}[c]{@{}c@{}}47.1~\cite{c42}, 56.0~\cite{c43},\\ 61.2~\cite{c44}, 64.2~\cite{c4}\end{tabular}                                 & 67.52                                                                            \\ \hline
		\multicolumn{1}{|l|}{MMI}   & \begin{tabular}[c]{@{}c@{}}51.4~\cite{c42}, 50.8~\cite{c2},\\ 36.8~\cite{c43}, 55.6~\cite{c4},\\ 66.9~\cite{c44}\end{tabular}               & 54.76                                                                            \\ \hline
		\multicolumn{1}{|l|}{FERA}  & \begin{tabular}[c]{@{}c@{}} 39.4\cite{c4} \end{tabular}                                                                                     & 41.93                                                                            \\ \hline
		\multicolumn{1}{|l|}{DISFA} & 37.7~\cite{c4}                                                                                                                              & 40.51                                                                                \\ \hline
	\end{tabular}
	\end{center}
	\caption{Recognition rates (\%) in cross-database task}
	\label{cross_database_table}
\end{table}

\begin{figure*}[!tbp]
	\centering
	\subfloat[CK+]{\includegraphics[width=.30\textwidth]{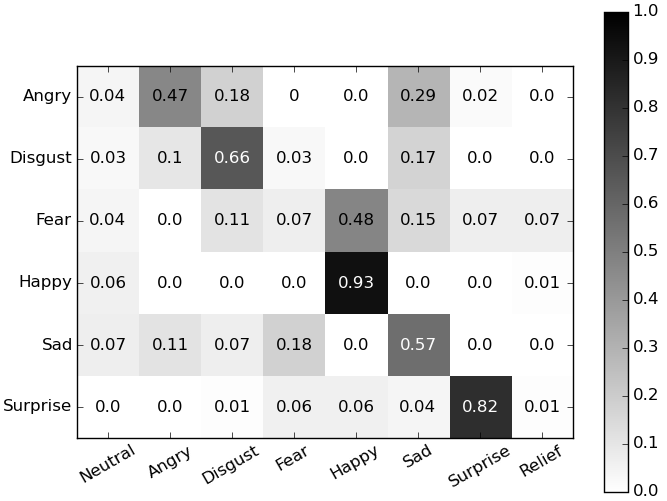}\label{fig:CK+_CM_cross_database}}
	\hspace{2em}
	\subfloat[MMI]{\includegraphics[width=.30\textwidth]{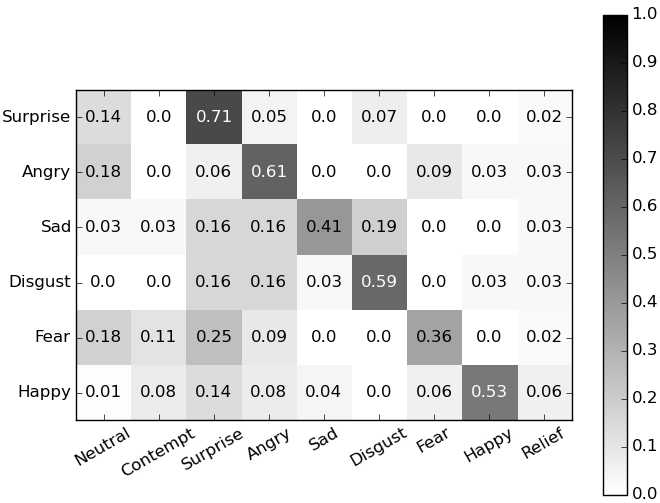}\label{fig:MMI_CM_cross_database}}
	\hspace{2em}\\
	\subfloat[FERA]{\includegraphics[width=.30\textwidth]{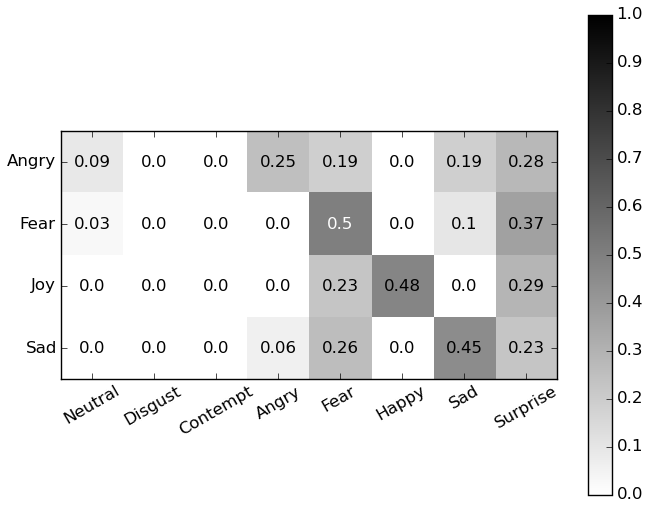}\label{fig:FERA_CM_cross_database}}
	\hspace{2em}
	\subfloat[DISFA]{\includegraphics[width=.30\textwidth]{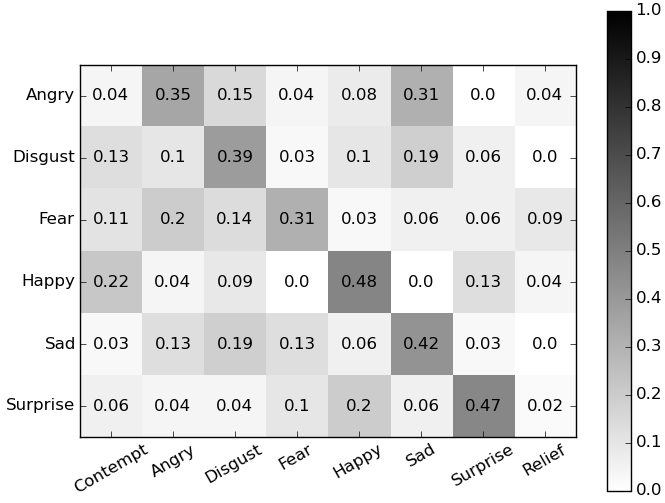}\label{fig:DISFA_CM_cross_database}}
	\caption{Confusion matrices of 3D Inception-ResNet with landmarks for cross-database task}
	\label{fig:confusion_matrices_cross_database}
\end{figure*}

In order to have a fair comparison with other methods, we provide the different settings used by the works mentioned in Table~\ref{cross_database_table}. The results provided in \cite{c42} are achieved by training the models on one of the CK+, MMI, and FEEDTUM databases and tested on the rest. The reported result in  \cite{c2} is the best achieved results using different SVM kernels trained on  CK+  and tested  on MMI database. In \cite{c43} several experiments were performed using four classifiers (SVM, Nearest Mean Classifier, Weighted Template Matching, and K-nearest neighbors). The reported results in this work for CK+ is trained on MMI and Jaffe databases while  the reported results for MMI is trained on the CK+ database only. As mentioned earlier, in \cite{c44} a Multiple Kernel Learning algorithm is used and the cross-database experiments are trained on CK+, evaluated on MMI and vice versa.  In \cite{c4} a DNN network is proposed using traditional Inception layer. The networks for the cross-database case in this work are tested on either CK+, MultiPIE, MMI, DISFA, FERA, SFEW, or FER2013 while trained on the rest. Some of the expressions of these databases are excluded in this study (such as neutral, relief, and contempt). There are other works that perform their experiments on action unit recognition task~\cite{chu2016modeling,fabian2016emotionet,jaiswal2016deep} but since fair comparison of action unit recognition and facial expression recognition is not easily obtainable, we did not mention these works in Tables~\ref{subject_ind_table} and~\ref{cross_database_table}.

Figure~\ref{fig:confusion_matrices_cross_database} shows the resulting confusion matrices of our experiments on 3D Inception-ResNet with landmarks in cross-database task. For CK+ (Figure~\ref{fig:CK+_CM_cross_database}), we exclude the contempt sequences in the test phase since other databases that are used for training the network, do not contain contempt category. Except for the fear expression (which has very few number of samples in other databases), the network has been able to correctly recognize other expressions. For MMI (Figure~\ref{fig:MMI_CM_cross_database}), highest recognition rate belongs to surprise while the lowest one belongs to fear. Also, we can see high confusion rate in recognizing sadness. On FERA (Figure~\ref{fig:FERA_CM_cross_database}), we exclude relief category as other databases do not contain this emotion. Considering the fact that only half of the train categories exist in the test set, the network shows acceptable performance in correctly recognizing emotions. However, surprise category has made significant confusion in all of the categories. On DISFA (Figure~\ref{fig:DISFA_CM_cross_database}), we exclude the neutral category as other databases do not contain this category. Highest recognition rates belong to happy and surprise emotions while lowest one belongs to fear. Comparing to other databases, we can see a significant increase in confusion rate in all of the categories. This can be in part due to the fact that emotions in DISFA are ``spontaneous" while emotions in the training databases are ``posed". Based on the aforementioned results, our method provides a comprehensive solution that can generalize well to practical applications.


\section{Conclusion}\label{sec:6}
In this paper, we presented a 3D Deep Neural Network for the task of facial expression recognition in videos. We proposed the 3D Inception-ResNet (3DIR) network which extends the well-known 2D Inception-ResNet module for processing image sequences. This additional dimension will result in a volume of feature maps and will extract the spatial relations between frames in a sequence. This module is followed by an LSTM which takes these temporal relations into account and uses this information to classify the sequences. In order to differentiate between facial components and other parts of the face, we incorporated facial landmarks in our proposed method. These landmarks are multiplied with the input tensor in the residual module which is replaced with the shortcuts in the traditional residual layer. 

We evaluated our proposed method in subject-independent and cross-database tasks. Four well-known databases were used to evaluate the method: CK+, MMI, FERA, and DISFA.  
Our experiments show that the proposed method outperforms many of the state-of-the-art methods in both tasks and provides a general solution for the task of FER.

\section{Acknowledgement}

This work is partially supported by the NSF grants IIS-1111568 and CNS-1427872. 
We gratefully acknowledge the support from NVIDIA Corporation with the donation of the Tesla K40 GPUs used for this research.


{\small
\bibliographystyle{ieee}
\bibliography{egbib}
}

\end{document}